\documentclass[english]{cccconf}
\usepackage[comma,numbers,square,sort&compress]{natbib}
\usepackage{epstopdf}
\usepackage{amsmath}
\usepackage{flushend}
\usepackage{multirow}
\usepackage{amssymb}

\begin{document}

\title{Dynamic Facial Expressions Analysis Based Parkinson's Disease Auxiliary Diagnosis}

\author{Xiaochen Huang\aref{1,2},
        Xiaochen Bi\aref{3},
        Cuihua Lv\aref{1,2},
        Xin Wang\aref{1,2},
        Haoyan Zhang\aref{4,}\textsuperscript{*},
        Wenjing Jiang\aref{3,}\textsuperscript{*},
        Xin Ma\aref{1,2,}\textsuperscript{*},
        Yibin Li\aref{1,2}}



\affiliation[1]{School of Control Science and Engineering, Shandong University, Jinan, 250061, China
        \email{huangxiaochen@mail.sdu.edu.cn, maxin@sdu.edu.cn}}
\affiliation[2]{Engineering Research Center of Intelligent Unmanned System, Ministry of Education, China
        \email{lvcuihua@mail.sdu.edu.cn, 202234930@mail.sdu.edu.cn}}
\affiliation[3]{Department of Geriatric Neurology, Qilu Hospital of Shandong University, Jinan, 250061, China
        \email{jiangwenjing@qiluhospital.com, bixiaochan406@126.com}}
\affiliation[4]{Shandong Inspur Science Research Institute Co., Ltd.
        \email{zhanghaoyan@inspur.com}}

\maketitle

\begin{abstract}
Parkinson's disease (PD), a prevalent neurodegenerative disorder, significantly affects patients' daily functioning and social interactions. To facilitate a more efficient and accessible diagnostic approach for PD, we propose a dynamic facial expression analysis-based PD auxiliary diagnosis method. This method targets hypomimia, a characteristic clinical symptom of PD, by analyzing two manifestations: reduced facial expressivity and facial rigidity, thereby facilitating the diagnosis process. We develop a multimodal facial expression analysis network to extract expression intensity features during patients' performance of various facial expressions. This network leverages the CLIP architecture to integrate visual and textual features while preserving the temporal dynamics of facial expressions. Subsequently, the expression intensity features are processed and input into an LSTM-based classification network for PD diagnosis. Our method achieves an accuracy of 93.1\%, outperforming other in-vitro PD diagnostic approaches. This technique offers a more convenient detection method for potential PD patients, improving their diagnostic experience.
\end{abstract}


\footnotetext{
Xiaochen Huang, Xiaochan Bi---Authors with equal contributions.

National Key Research and Development Program Project under Grant 2023YFB4706104.
Key R\&D Program of Shandong Province (Major Science and Technology Innovation Project) under Grant 2024CXGC010603.
Fundamental Research Funds for the Central Universities under Grant 2022JC011.
}

\section{Introduction}

Parkinson's disease (PD), a highly prevalent neurodegenerative disorder, significantly affects the daily functioning and social interactions of middle-aged and elderly individuals \cite{balestrino2020parkinson}. Common clinical symptoms of PD include tremors, freezing of gait, dysarthria, and Hypomimia \cite{bloem2021parkinson}. Current diagnostic methods for PD primarily rely on clinical scales, such as the Unified Parkinson's Disease Rating Scale (UPDRS), Non-Motor Symptoms Scale (NMSS), and Non-Motor Symptoms Questionnaire (NMSQuest) \cite{movement2003unified, chaudhuri2007metric, romenets2012validation}. However, scale-based diagnosis is heavily dependent on the clinician’s expertise, introducing a margin for error. Additionally, there is a shortage of specialized neurologists in remote areas, further complicating accurate diagnosis. Thus, there is an urgent need for an auxiliary diagnostic method for PD that offers both high accuracy and convenience.

Existing in-vitro auxiliary diagnostic methods based on the typical symptoms of PD enable patients to achieve an initial diagnosis at home with high accuracy \cite{rohr2016value}. These methods diagnose PD by analyzing gait signals \cite{su2021simple, wu2023advantages}, voice signals \cite{ahmed2022classification, karaman2021robust}, or facial expression signals \cite{oliveira2023tabular, abrami2021automated}. However, gait signal acquisition requires the use of motion sensors, and the reliability of voice signal analysis can be affected by regional variations in speaking habits. In contrast, facial expression signals can be easily acquired using a smartphone, and Hypomimia is a characteristic clinical symptom in most PD patients. Therefore, facial expression-based diagnostic methods for PD present significant potential for development.

Patients with Hypomimia often exhibit both facial expression absence and facial rigidity \cite{ricciardi2020hypomimia}. Most existing facial-based diagnostic methods for PD primarily analyze static facial images, failing to account for the symptom of facial rigidity. To address this limitation, we employ dynamic facial expression analysis to evaluate the facial expressions of PD patients by assessing videos of patients performing various facial expressions \cite{fasel2003automatic}. This approach allows us to quantify the intensity of the performed expressions and ultimately facilitates the diagnosis of PD:

\begin{itemize}
    \item We employ a multimodal dynamic facial expression analysis network founded on the CLIP architecture to facilitate the extraction of expression intensity. In comparison to single-modal input networks, the multimodal network is capable of learning shared features between images and text, thereby demonstrating enhanced generalization capabilities and achieving superior performance in transfer learning.
    \item We develop a dynamic facial expression analysis based PD auxiliary diagnosis network. The dynamic expression analysis network evaluates videos of patients performing various facial expressions. The resulting expression intensity features reflect both the symptoms of facial expression absence and facial rigidity, enabling a more comprehensive analysis of hypomimia.
\end{itemize}

The remainder of this paper is organized as follows. Section 2 presents the Methodology. Section 3 shows the experimental and results. Section 4 presents the discussion and conclusion. 

\begin{figure*}[!t]
    \centering
    \includegraphics[width=1\linewidth]{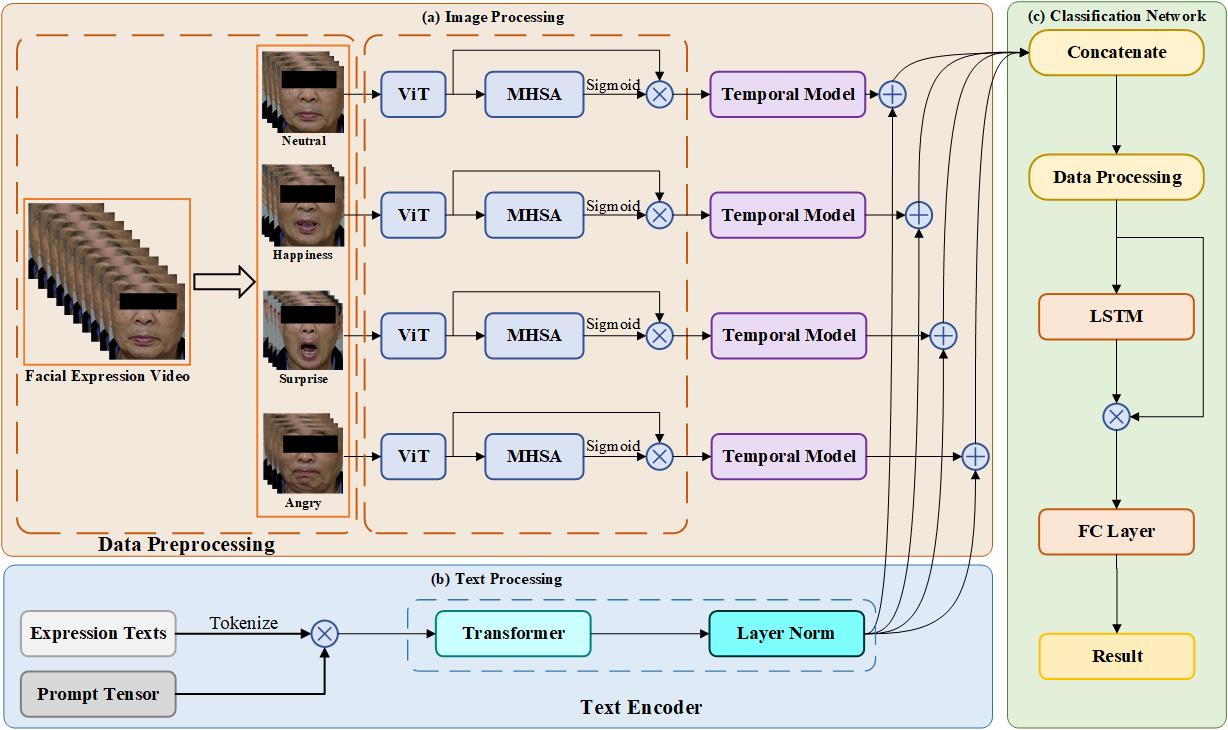}
    \caption{Structure of Dynamic Facial Expressions Analysis Based Parkinson's Disease Auxiliary Diagnosis Network.}
    \label{fig1}
\end{figure*}

\section{Methods}

This section delineates the architecture of the proposed dynamic facial expression analysis based PD auxiliary diagnosis network. Initially, we design a multimodal dynamic expression analysis network to extract the intensity of patients' facial expressions. This model incorporates the CLIP architecture to simultaneously analyze both visual and textual features, thereby enhancing accuracy \cite{hafner2021clip}. Subsequently, the extracted expression intensity features are processed to further enhance the highlight values before being input into an LSTM-based classification network for PD auxiliary diagnosis. Furthermore, four fundamental expressions—neutral, happiness, surprised, and angry—are experimentally selected as inputs to the network. This selection simplifies the training process while ensuring the accuracy of the diagnosis. The network structure is illustrated in Fig.~\ref{fig1}.

\subsection{Datasets}

This subsection outlines the dataset employed in the training of dynamic facial expression analysis based PD auxiliary diagnostic network.

\subsubsection{Dynamic Facial Expression in-the-Wild (DFEW) Dataset}

DFEW is a large-scale facial expression database comprising 16,372 challenging video clips sourced from films \cite{jiang2020dfew}. For training the dynamic expression analysis model, we select video data representing four specific emotions: neutral, happiness, surprised, and angry. The composition of the dataset utilized for training is illustrated in Table ~\ref{tab1}.

\begin{table}[h]
    \centering
    \caption{Composition of the DFEW Dataset.}
    \begin{tabular}{ccccc}
        \hline
             & Neutral & happiness & surprised & angry\\
        \hline
            Training Sets & 2196 & 2017 & 1235 & 1799\\
            Testing Sets & 549 & 504 & 310 & 450\\
        \hline
    \end{tabular}
    \label{tab1}
\end{table}

\subsubsection{Parkinson’s Disease Facial Expression Videos (PD-FEV) Dataset}

In collaboration with Qilu Hospital of Shandong University, we collect video data of PD patients continuously performing seven fundamental facial expressions to address the deficiency of existing facial expression datasets for PD patients. During data collection, patients are instructed to perform each facial expression for a duration exceeding five seconds. To optimize participant performance, verbal guidance is provided throughout the performance process, and supplemental lighting from a lamp source is utilized to enhance illumination on the participants' faces. PD-FEV dataset comprises video recordings of facial expressions from a total of 173 individuals, which includes 97 individuals diagnosed with PD and 76 healthy controls (HC). The participant demographics of PD-FEV dataset are shown in Table ~\ref{tab3}.

\begin{table}[h]
    \centering
    \caption{Participant demographics of PD-FEV dataset.}
    \begin{tabular}{p{2.19cm}p{2.19cm}p{2.19cm}}
    \hline
        Variable & PD & HC \\
    \hline
        Number & 97 & 76\\
        Sex(M/F) & 46/51 & 22/54\\
    \hline
    \end{tabular}
    \label{tab3}
\end{table}

\subsection{Dynamic Facial Expressions Analysis Network}

A CLIP-based dynamic expression recognition network is devoloped to extract facial expression intensity features from PD patients as they perform various facial expressions. This network consists of two components: a visual part and a textual part. In the visual part, we incorporate a Multi-Head Self-Attention (MHSA) module into the CLIP image encoder to enhance sequence processing. Additionally, we introduce a temporal model based on the Transformer encoder to preserve the temporal features of continuous expression videos. In the textual part, we employ descriptive phrases of facial expressions, rather than merely using class labels, as input text. Furthermore, we integrate prompt tensors to bolster the model's capacity to capture contextual information within the text. The features extracted from both the visual and textual parts are combined to derive the intensity of the patients' expressions. We will present the visual and textual components separately.

\subsubsection{Visual Part}

We sample \(M\) frames from the facial expression videos, after which an image encoder is constructed to extract features from these sampled images. The image encoder comprises two modules: the Vision Transformer (ViT) module ${{E}_{1}}(\cdot )$ and the Multi-Head Self-Attention (MHSA) module ${{E}_{2}}(\cdot )$, which collectively extract image features ${{f}_{i}}$ for each sampled frame ${{m}_{i}}$ where $i\in \{1,2,\cdots ,M\}$. Subsequently, the \(M\) feature vectors are input into the temporal model ${T}(\cdot )$, from which the final visual features \(F\) are derived. This process can be expressed as follows:

\begin{equation}
  \label{eq1}
    \begin{aligned}
        & {{A}_{i}}={{E}_{1}}({{m}_{i}}) \\ 
        & {{{\tilde{A}}}_{i}}={{E}_{2}}({{A}_{i}}) \\ 
        & {{f}_{i}}={{A}_{i}}*Sigmoid({{{\tilde{A}}}_{i}}) \\ 
        & F=T({{f}_{1}},{{f}_{2}},\ldots ,{{f}_{M}})
    \end{aligned}
\end{equation}

\subsubsection{Textual Part}

For the textual part, the input textual data is first subjected to Tokenization, converting each segment of text into Prompts \textit{T} to facilitate the model's subsequent understanding and processing. During this process, Tokenization not only considers the segmentation of words or subwords but also ensures that the text is transformed into a standardized form for subsequent computations. Additionally, this study constructs Tokenized Prompts $\tilde{T}$ for each text segment. These prompt vectors incorporate contextual information to assist the model in more accurately interpreting the semantic structure and intent of the text.

Overall, we initially utilize descriptive phrases corresponding to four expressions—neutral, happiness, surprised, and angry—as input text. We use textual descriptions of different facial expressions as input to extract the corresponding prompts \textit{T} and Tokenized Prompts $\tilde{T}$. These vectors are then input into the text encoder ${M}(\cdot )$ to obtain the final text features \textit{G}. This process can be represented as:

\begin{equation}
  \label{eq2}
   G=M(T,\tilde{T})
\end{equation}

Subsequently, the visual features \(F\) and textual features \(G\) are integrated to derive the intensity of the patient's facial expression. The integration process is defined as:

\begin{equation}
  \label{eq3}
   I=\exp (\cos (F,G)/\tau )
\end{equation}

Where $\cos (\cdot ,\cdot )$  denotes cosine similarity and $\tau $ is a hyperparameter.

\subsection{Data Processing and Classification Network}

Following the dynamic expression analysis network, four expression intensity features are extracted for each expression video, resulting in a total of sixteen bits of expression intensity feature $I=\left\{ {{i}_{1}},{{i}_{2}},\cdots ,{{i}_{16}} \right\}$ for each patient after inputting four expression videos. To highlight the expression intensity features corresponding to the same expression labels as the input video, this study processes the expression intensity features \textit{I}. During the data processing, the expression intensity features \textit{I} is first divided into four groups, namely ${{I}_{j}}=\left\{ {{i}_{4(j-1)+1}},{{i}_{4(j-1)+2}},{{i}_{4(j-1)+3}},{{i}_{4j}} \right\}$, $j\in \{1,2,3,4\}$, based on the input facial expression videos.

During data processing we calculate he Highlight Value (\textit{hv}), Mean, Standard Deviation (\textit{std}), Z-Score, Percentage Difference (\textit{pd}), Range, Difference from Minimum (${D}_{min}$), and Difference from Maximum (${D}_{max}$), with their corresponding formulas defined as follows: 

\begin{equation}
  \label{eq4}
    \begin{aligned}
        &     hv=\left\{ \begin{matrix}
       {{i}_{1}},j=1  \\
       {{i}_{6}},j=2  \\
       {{i}_{11}},j=3  \\
       {{i}_{16}},j=4  \\
    \end{matrix} \right. \\
        & Mean=\frac{1}{4}\sum\limits_{k=1}^{4}{{{i}_{4(j-1)+k}}} \\
        & std=\sqrt{\frac{1}{4}\sum\limits_{k=1}^{4}{{{({{i}_{4(j-1)+k}}-Mean)}^{2}}}} \\
        & Z-Score=\frac{\textit{hv}-Mean}{std} \\
        & pd=\frac{\textit{hv}-Mean}{Mean} \\
        & Range=\max ({{I}_{j}})-\min ({{I}_{j}}) \\
        & {{D}_{\min }}=\textit{HV}-\min ({{I}_{j}})\\
        & {{D}_{\max }}=\textit{HV}-\max ({{I}_{j}})
    \end{aligned}
\end{equation}

The \textit{std} measures the degree of data dispersion, reflecting the deviation of data points from the Mean and revealing the overall variability of the data. The Z-Score and \textit{pd} indicate the difference between the Highlight Value and the Mean, which corresponds to the distance between the labeled expression intensity and the mean intensity. The Range represents the difference between the maximum and minimum values within the group, reflecting the overall extent of variation and highlighting the difference in expression intensity extremes within the video. The ${D}_{min}$ and ${D}_{max}$ quantify the deviation of the Highlight Value from the group’s minimum or maximum, revealing the position of the Highlight Value within the group.

Upon completing the data processing, we conduct the classification of PD using an LSTM-based classification network. The accuracy of the classification network is enhanced by incorporating a residual structure within the LSTM framework, thereby mitigating issues related to gradient vanishing and performance degradation \cite{cox1968general, sherstinsky2020fundamentals}. The achieved accuracy following five-fold cross-validation is 93.1\%.

\begin{figure*}[!t]
    \centering
    \includegraphics[width=0.7\linewidth]{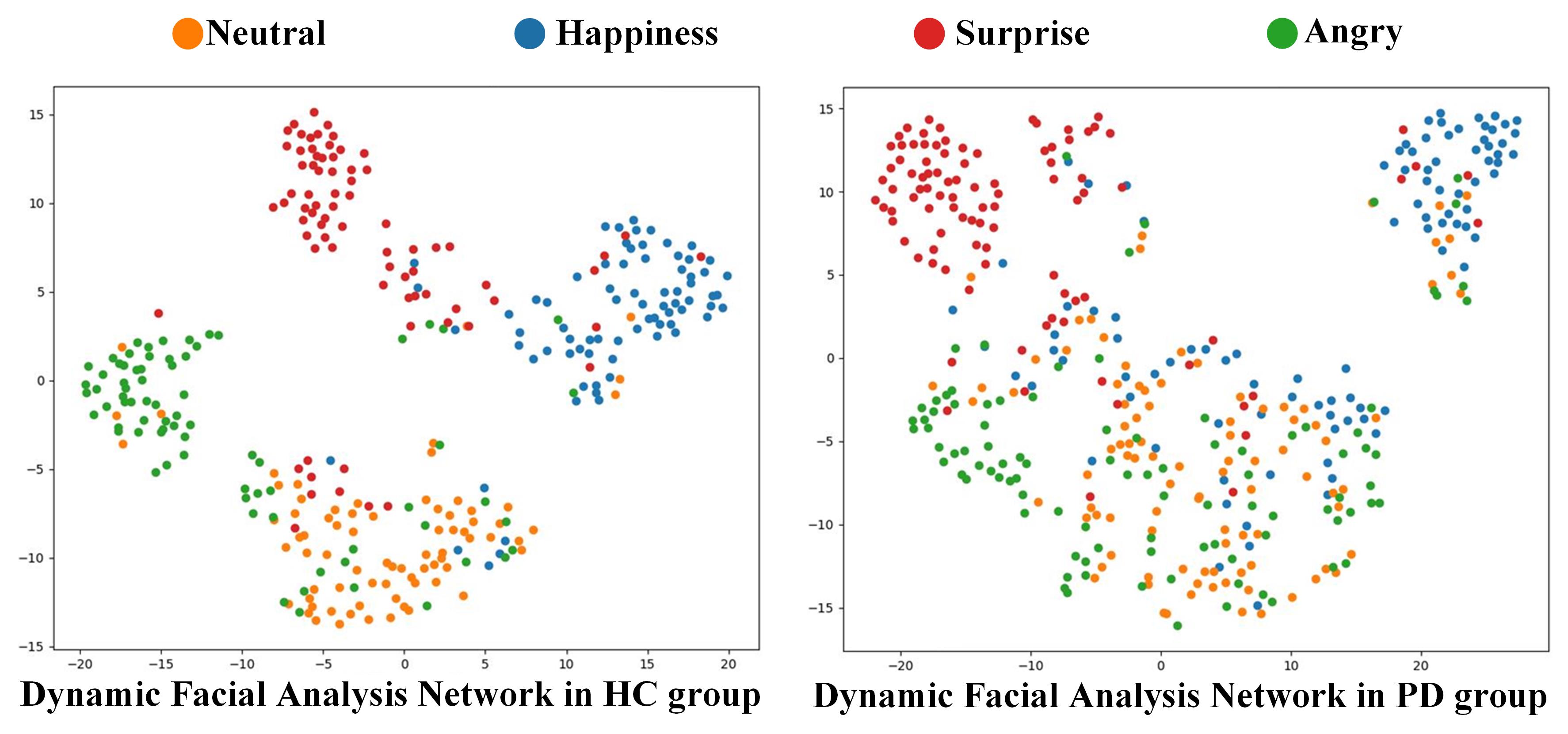}
    \caption{Visualization results of Dynamic Facial Analysis Network in Different Groups.}
    \label{fig5}
\end{figure*}

\section{Experiments and Results}

In this section, we will provide a detailed account of the experimental design for dynamic expression analysis network, as well as the setup and the results of ablation study.

\subsection{Design of Dynamic Facial Expression Analysis Experiments}

\begin{figure} [!h]
    \centering
    \includegraphics[width=0.9\linewidth]{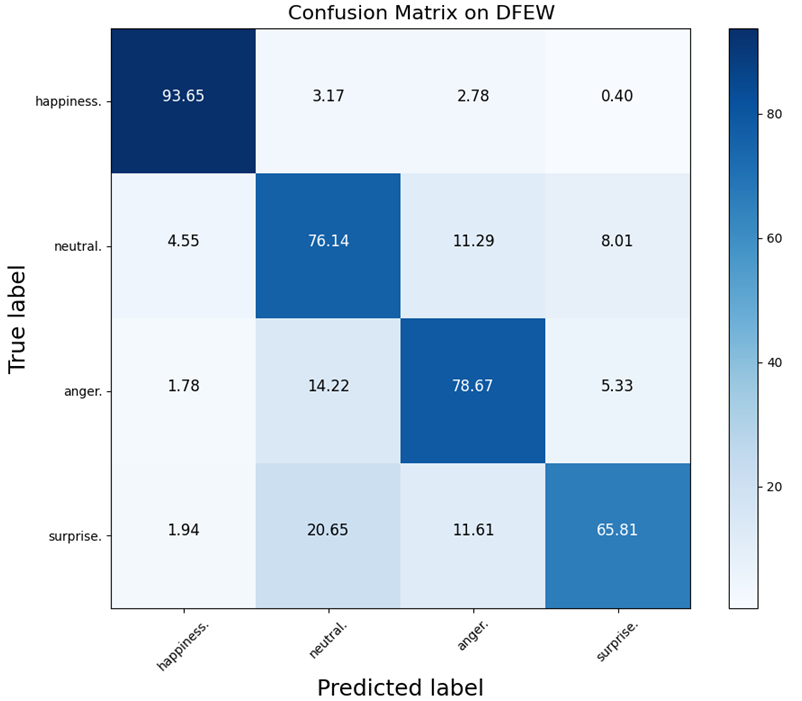}
    \caption{Confusion matrix of dynamic facial expression analysis network.}
    \label{fig2}
\end{figure}

The DFER dataset is selected to train the dynamic facial expression analysis network. The network is trained with a batch size of 16 over 50 epochs. The learning rate for the image encoder is set at $1\times {{10}^{-5}}$, while the learning rate for the temporal model is set at $1\times {{10}^{-3}}$. The proposed model is implemented using PyTorch, and all experiments are conducted on a workstation equipped with a 13th Gen Intel(R) Core(TM) i9-13900K CPU and an NVIDIA GeForce RTX 4090 GPU. The confusion matrix is presented in Figure ~\ref{fig2}.

We subsequently evaluate the trained expression recognition network on the PD-FEV dataset, achieving an accuracy of 80.9\% for the HC group and 58.2\% for the PD patient group. These results further indicate that PD patients exhibit a diminished capacity to perform expressions compared to the healthy counterparts. Figure ~\ref{fig5} presents the t-SNE visualization results of the dynamic facial expression analysis network for the PD and HC groups. In the visual analysis of the PD group, although surprise and happiness expressions still show a certain level of distinction, indicating that patients are able to perform facial muscle coordination to some extent, the classification performance significantly declines compared to the healthy control group. In classification tasks involving neutral and angry expressions, effective differentiation is almost impossible. This phenomenon reflects the significant limitation in facial expression expression in PD patients, primarily caused by factors such as facial muscle rigidity, bradykinesia, and a reduction in involuntary facial movements, leading to substantial difficulty when mimicking complex expressions or those requiring greater facial muscle movement.

\begin{figure*} [!t]
    \centering
    \includegraphics[width=0.9\linewidth]{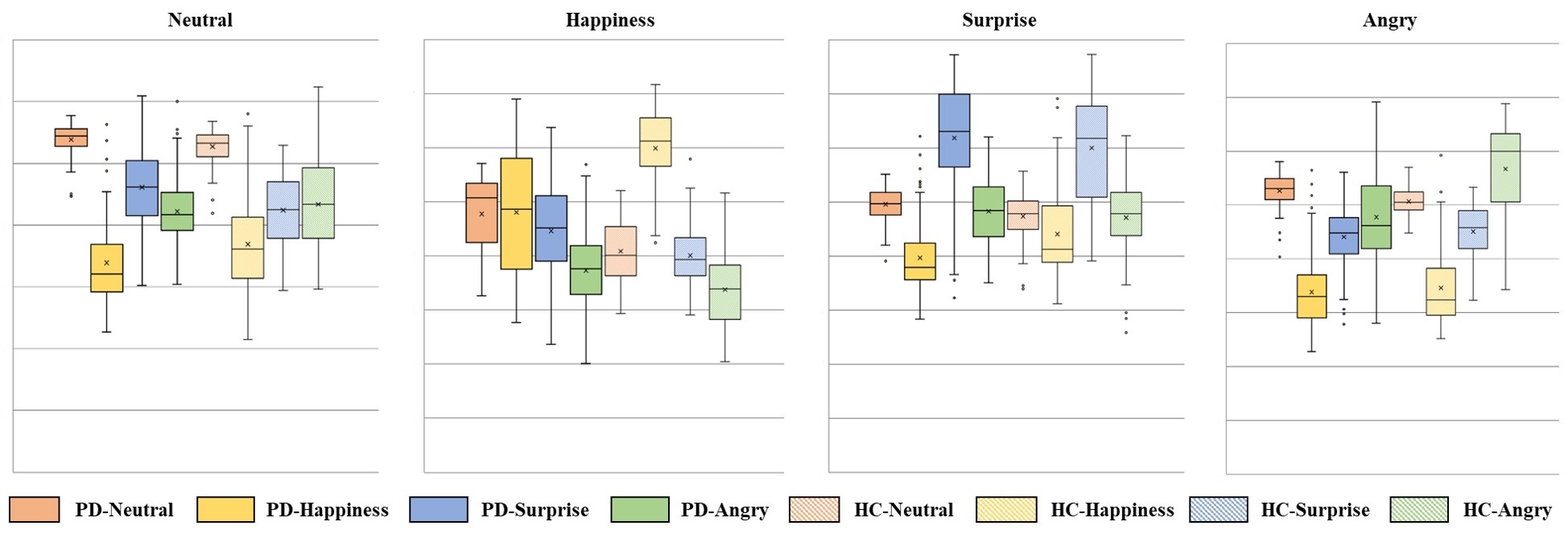}
    \caption{Boxplots Depicting the Varying Intensities of Different Facial Expressions in Each Video.}
    \label{fig3}
\end{figure*}

Boxplots depicting the varying intensities of different facial expressions in each video are presented in Figure ~\ref{fig3}. The analysis reveals a significant difference in the intensity of happy expressions between PD patients and the HC group in the happy expression videos, aligning with the findings from clinical medical trials. The next most pronounced difference is observed in the angry expression, indicating that PD patients exhibit a diminished ability to frown compared to the HC group. Conversely, the neutral expression demonstrates the lowest degree of distinction, suggesting that the differences between PD patients and the HC group are minimal in the absence of facial muscle movement.

\begin{table*}[h]
    \centering
    \caption{Results of Ablation Experiment.}
        \begin{tabular}{ccccccc}
            \hline
            Classification Network & Data Processing & Residual & Accuracy & Precision & Recall & F1-Score\\
            \hline
            \multirow{2}{*}{\centering DT} & - & - & 0.633 & 0.576 & 0.655 & 0.611\\
             & \checkmark & - & 0.663 & 0.622 & 0.591 & 0.605\\
             \hline
             \multirow{2}{*}{\centering BC} & - & - & 0.767 & 0.721 & 0.782 & 0.744\\
             & \checkmark & - & 0.727 & 0.653 & 0.831 & 0.727\\
             \hline
             \multirow{2}{*}{\centering RF} & - & - & 0.738 & 0.704 & 0.678 & 0.689\\
             & \checkmark & - & 0.733 & 0.684 & 0.689 & 0.683\\
             \hline
             \multirow{2}{*}{\centering SVM} & - & - & 0.749 & 0.709 & 0.764 & 0.724\\
             & \checkmark & - & 0.758 & 0.720 & 0.773 & 0.738\\
             \hline
             \multirow{4}{*}{\centering GRU} & - & - & 0.715 & 0.717 & 0.715 & 0.714 \\
             & \checkmark & - & 0.750 & 0.764 & 0.750 & 0.747\\
             & - & \checkmark & 0.721 & 0.731 & 0.722 & 0.722\\
             & \checkmark & \checkmark & 0.743 & 0.755 & 0.743 & 0.742\\
             \hline
             \multirow{4}{*}{\centering \textbf{LSTM}} & - & - & 0.885 & 0.890 & 0.885 & 0.885\\
             & \checkmark & - & 0.908 & 0.912 & 0.908 & 0.906\\
             & - & \checkmark & 0.909 & 0.912 & 0.909 & 0.909\\
             & \checkmark & \checkmark & \textbf{0.931} & \textbf{0.929} & \textbf{0.931} & \textbf{0.929}\\
            \hline
        \end{tabular}
    \label{tab2}
\end{table*}

\subsection{Ablation Experiment}

To assess the necessity of data processing and the rationale behind the structural design of the LSTM-based classification network, we conduct the ablation experiments.

\subsubsection{Experiment Setup}

To validate the efficacy of the classification network, we select various classification networks and structural configurations for ablation experiments, conducting evaluations on both the original and processed data using the same network architectures. In addition to LSTM, we employ Bayesian Classifiers (BC), Random Forests (RF), Decision Trees (DT), Support Vector Machines (SVM) and Gated Recurrent Units (GRU) to assess and compare their performance \cite{breiman2001random, rokach2005decision, hearst1998support}.

\subsubsection{Experiment Results}

As demonstrated in Table 2, simpler models such as DT and VM do not exhibit a substantial improvement in classification performance after the features are processed. However, in more complex models like GRU and LSTM, the facial expression features extracted by the data processing significantly enhanced classification capabilities. In the LSTM-based classification network used in this study, the combination of processed features led to improvements of 2.2\%, 1.7\%, 2.2\%, and 2.0\% in accuracy, precision, recall, and F1-Score, respectively. These results highlight that the data processing significantly improved the performance of the LSTM-based network, validating its effectiveness in the proposed dynamic facial expression analysis network.

The table also compares the performance of different classification networks on the same dataset. The LSTM-based network outperformed others, showing stronger classification performance by effectively capturing temporal features in sequential data. This allowed for improved accuracy, precision, recall, and F1-Score. Thus, the LSTM-based network demonstrates higher classification accuracy, especially in dynamic emotional expression scenarios.

\section{Conclusion and Disscussion}

Hypomimia, recognized as a characteristic clinical symptom of Parkinson's disease (PD), primarily manifests as a facial expression absence and facial rigidity in affected individuals. Consequently, an auxiliary diagnostic method for PD can be developed based on the detection of these two features. Given that hypomimia typically presents in the early stages of the disease, an auxiliary diagnostic approach targeting this symptom can facilitate earlier detection and prompt initiation of treatment, potentially slowing disease progression. In this study, we propose an auxiliary diagnostic method for PD based on the analysis of dynamic facial expressions. This method involves evaluating videos of patients performing various expressions to extract the intensity of facial expressions, while concurrently addressing both the absence of facial expression and facial rigidity, thereby enabling a more comprehensive auxiliary diagnosis for hypomimia.

We extract dynamic expression intensity features using a multimodal dynamic expression analysis network based on the CLIP architecture. This network integrates both visual and textual information, enabling the learning of shared features between images and text, while employing a temporal model to preserve the temporal characteristics of the original input data. Subsequently, we conduct data processing on the facial expression intensity features to enhance its salient features and construct an LSTM-based classification network, aimed at achieving improved diagnostic accuracy.

Despite these encouraging results, our research is not without limitations. In the following subsections, we will discuss these limitations and the future research outlook.

\subsection{Limitations}

The PD-FEV dataset utilized for training the network comprise facial video data collected in a controlled laboratory environment, where we ensure optimal lighting conditions and provide verbal guidance to participants to enhance performance. However, not all subjects seeking assisted diagnosis have their facial videos captured under such favorable lighting conditions, which adversely affect the classification accuracy of the network. Furthermore, our dynamic facial expression analysis network features a substantial number of parameters and a complex network architecture, presenting challenges for deployment on lightweight clients for training and application. Addressing these issues will be a critical focus in future research endeavors.

\subsection{Future Work}

In future work, we aim to expand the PD-FEV dataset by collecting video data of facial expressions across various environments. This will enhance the PD assisted diagnosis network's classification capability for facial images captured in diverse settings, thereby improving the network's robustness. Additionally, we will explore more lightweight models for PD diagnosis to reduce model complexity while maintaining accuracy. This approach aims to facilitate deployment on lightweight devices, enabling diagnosis for a broader range of PD patients who may benefit from such assistance.

\section*{Declaration of Interest Statement}
The authors declare that they have no known competing financial interests or personal relationships that could have appeared to influence the work reported in this paper.

\section*{Data availability}
The data that has been used is confidential.

\bibliographystyle{unsrt}

\bibliography{cas-refs}

\end{document}